\title{Premise arXiv (v2)}
\documentclass[11pt,letterpaper]{article}
\usepackage{emnlp2017}
\usepackage{times}
\usepackage{latexsym}

\usepackage{fancyhdr}

\usepackage{relsize}
\usepackage[T1]{fontenc} 
\usepackage[latin1]{inputenc} 
\usepackage[english]{babel}
\usepackage{ragged2e}

\usepackage{hyperref}

\usepackage{epsfig}
\usepackage{graphicx}
\usepackage{wrapfig}
\usepackage[belowskip=-7pt,aboveskip=5pt,font=small]{caption}
\usepackage[belowskip=-2pt,aboveskip=0pt,font=small]{subcaption}
\usepackage{amsmath, amsthm, amssymb}
\usepackage{textcomp}
\usepackage{stmaryrd}
\usepackage{upgreek}
\usepackage{bm}
\usepackage{cases}
\usepackage{mathtools}

\usepackage{algorithm}
\usepackage{algpseudocode}

\usepackage{multirow}
\usepackage{rotating}
\usepackage{booktabs}
\usepackage{tabularx}

\usepackage{enumitem}
\usepackage[olditem,oldenum]{paralist}

\usepackage{alltt}
\usepackage{listings}
\usepackage{url}
\usepackage{xspace}
\usepackage{comment}
\usepackage{color}
\usepackage{afterpage}
\usepackage{framed}
\usepackage{fancybox}
\usepackage[normalem]{ulem}
\usepackage{pdfcomment}

\usepackage{soul}
\usepackage{color}
\usepackage{xargs}  
\usepackage{dblfloatfix}
\belowdisplayskip \abovedisplayskip

\newlength{\sectionReduceTop}
\newlength{\sectionReduceBot}
\newlength{\subsectionReduceTop}
\newlength{\subsectionReduceBot}
\newlength{\abstractReduceTop}
\newlength{\abstractReduceBot}
\newlength{\captionReduceTop}
\newlength{\captionReduceBot}
\newlength{\subsubsectionReduceTop}
\newlength{\subsubsectionReduceBot}

\newlength{\eqnReduceTop}
\newlength{\eqnReduceBot}

\newlength{\horSkip}
\newlength{\verSkip}

\newlength{\figureHeight}
\usepackage{mysymbols}

\newcommand{\tabref}[1]{Table \ref{#1}}

\newcommand{\sml}[1]{\textcolor{black}{#1}}

\emnlpfinalcopy

\def\emnlppaperid{7}

\title{The Promise of Premise: Harnessing Question Premises\\ in Visual Question Answering}

\author{Aroma Mahendru\thanks{\,\,\,Denotes equal contribution.}$\,\,^{,1}$ \enskip Viraj Prabhu$^{*,1}$ \enskip Akrit Mohapatra$^{*,1}$ \enskip Dhruv Batra$^{2}$ \enskip Stefan Lee$^{1}$ \vspace{0.005\textwidth} \\
$^1$Virginia Tech \qquad $^2$Georgia Institute of Technology\\
\hspace{-2mm}{\tt\{maroma, virajp, akrit\}@vt.edu},\enskip {\tt dbatra@gatech.edu},\enskip {\tt steflee@vt.edu}}

\date{}

\begin{document}

\pagestyle{fancy}
\renewcommand{\headrulewidth}{0pt}
\fancyhf{}
\chead{\vspace{-50pt}\normalsize Published at the 2017 Conference on Empirical Methods on Natural Language Processing (EMNLP)\vspace{20pt}}

\maketitle
\vspace{\abstractReduceTop}
\begin{abstract}
	\vspace{\abstractReduceBot}
In this paper, we make a simple observation that questions about images often contain \emph{premises} -- objects and relationships implied by the question -- and that reasoning about premises can help Visual Question Answering (VQA) models respond more intelligently to irrelevant or previously unseen questions.

When presented with a question that is irrelevant to an image, state-of-the-art VQA models will still answer purely based on learned language biases, resulting in nonsensical or even misleading answers. We note that a visual question is irrelevant to an image if at least one of its premises is false (\ie not depicted in the image).  We leverage this observation to construct a dataset for Question Relevance Prediction and Explanation (QRPE) by searching for false premises. We train novel question relevance detection models and show that models that reason about premises consistently outperform models that do not.

We also find that forcing standard VQA models to reason about premises during training can lead to improvements on tasks requiring compositional reasoning.
\end{abstract}

\vspace{\sectionReduceTop}
\section{Introduction}
\label{sec:intro}
\vspace{\sectionReduceBot}

The task of providing natural language answers to free-form questions about an image -- \ie Visual Question Answering (VQA) -- 
has received substantial attention in the past few years \cite{malinowski2014multi,antol2015vqa,malinowski2015ask,zitnick2016measuring,kim2016multimodal,wu2016value,lu2016hierarchical,andreas2016neural,LuXPS16} and has quickly become a popular problem area. Despite significant progress on VQA benchmarks \cite{antol2015vqa}, 
current models still present a number of unintelligent and problematic tendencies. 

\begin{figure}[t]
	\centering
    \vspace{5pt}
	\includegraphics[width=\linewidth]{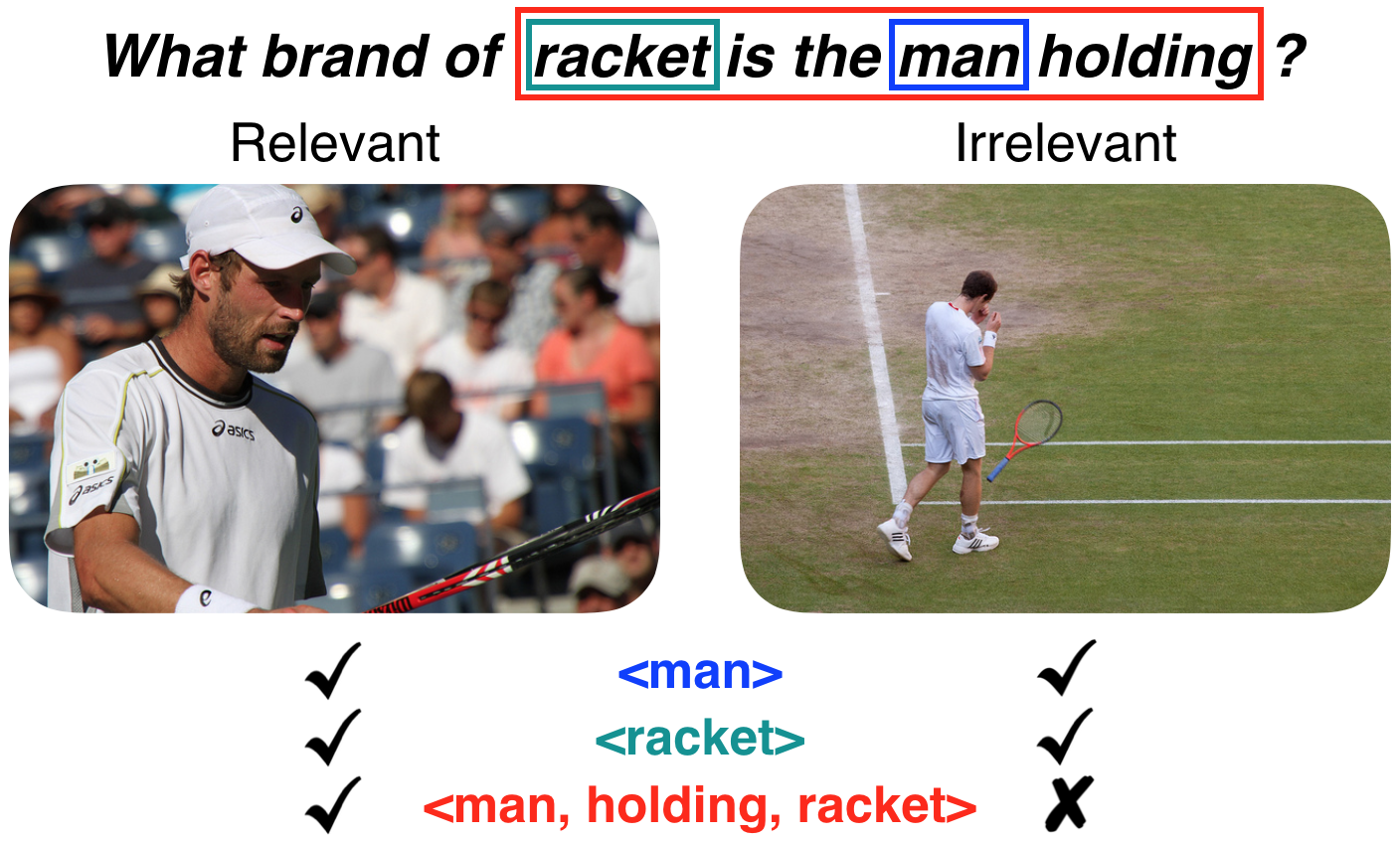}
	\caption{
 	Questions asked about images often contain \emph{`premises'} that imply visual semantics. From the  above question, we can infer that a relevant image must contain a man, a racket, and that the man must be holding the racket. We extract these premises from visually grounded questions and use them to construct a new dataset and models for question relevance prediction. We also find that augmenting standard VQA training with simple premise-based questions results in improvements on tasks requiring compositional reasoning.
    }
\label{fig:intro}
\end{figure}

When faced with questions that are irrelevant or not applicable for an image, current `forced choice' models will still produce an answer. For example, given an 
image of a dog and a query ``\textit{What color is the bird?}'', standard VQA models might answer ``\textit{Red}'' confidently, based 
solely on language biases in the training set (\ie an overabundance of the word ``red''). In these cases, the predicted answers are senseless at best and misleading at worst, 
with either case posing serious problems for real-world applications. Like \citet{ray2016question}, we argue that practical VQA systems must be able to identify and explain irrelevant questions. For instance, a more intelligent VQA model with this capability 
might answer ``\textit{There is no bird in the image}'' for this example.

\paragraph{Premises.} In this paper, we show that question \emph{premises} - \ie objects and relationships implied by a question - can enable VQA models to respond more intelligently to irrelevant or previously unseen questions. We develop a premise extraction pipeline based on SPICE~\cite{anderson2016spice} and demonstrate how these premises can be used to improve modern VQA models in the face of irrelevant or previously unseen questions. 

Concretely, we define premises as facts implied by the language of questions, for example the question
 \textit{``What brand of racket is the man holding?''} shown in \figref{fig:intro} implies the existence of a man, a racket, and that the man is holding the racket. For visually grounded questions (\ie those asked about a particular image) these premises 
imply visual qualities, including the presence of objects as well as their attributes and relationships.

Broadly speaking, we explore the usefulness of premises in two settings -- when visual questions are known to be relevant to the images they are asked on (\eg in the VQA dataset) and in real-life situations where such an assumption cannot be made (\eg when generated by visually impaired users). In the former case, we show that knowing that a question is relevant allows us to perform data augmentation by creating additional simple question-answer pairs using the premises of source questions.  
In the latter case, we show that explicitly reasoning about premises provides an effective and interpretable way of determining whether a question is relevant to an image.

\paragraph{Irrelevant Question Detection.} We consider a question to be relevant to an image if all of the question's premises apply to the corresponding image, that is to say all objects, 
attributes, and interactions implied by the question are depicted in the image. We refer to premises that apply for a given image as true premises and those that do not apply as false premises. In order to train and evaluate models for this task, we curate a new irrelevant question detection dataset which we call the Question Relevance Prediction and Explanation (QRPE) dataset. QRPE is automatically curated from annotations already present in existing datasets, requiring no additional labeling.

We collect the QRPE dataset by taking each image-question pair in the VQA dataset \cite{antol2015vqa} and finding the most 
visually similar other image for which exactly one of the question premises is false. In this way, we collect tuples consisting of two 
images, a question, and a premise where the question is relevant for one image and not for the other due to the premise being false. 

For context, the only other
existing irrelevant question detection dataset
\cite{ray2016question} collected irrelevant question-image pairs by human verification of random pairs. In comparison, QRPE  is substantially larger, balanced between irrelevant and relevant examples, and presents a 
considerably more difficult task due to the closeness of the image pairs both visually and with respect to question premises.

We train novel models for irrelevant question detection on the QRPE dataset and compare to existing methods. In these experiments,
we show that models that explicitly reason about question premises consistently outperform baseline models that do not. 

\paragraph{VQA Data Augmentation.} Finally, we also introduce an approach to generate simple, templated question-answer pairs about elementary concepts from premises of complex training questions. In initial experiments, we show that adding these simple question-answer pairs to VQA training data can improve performance
on tasks requiring compositional reasoning. These simple questions improve training by bringing implicit training concepts ``to the surface'', 
\ie introducing direct supervision of important implicit concepts by transforming them to simple training pairs.
\vspace{\sectionReduceTop}
\section{Related Work}
\label{sec:related}
\vspace{\sectionReduceBot}

\paragraph{Visual Question Answering:} Starting from simple bag-of-word and CNN+LSTM models
\cite{antol2015vqa}, VQA architectures have seen considerable innovation. Many top-performing models integrate
attention mechanisms (over the image, the question, or both) to focus on important structures
\cite{fukui2016multimodal,lu2016hierarchical,LuXPS16}, and some have been designed with compositionality in mind \cite{andreas2016neural,hendricks16cvpr}. However, improving compositionality or performance through data augmentation remains a largely unstudied area. 

Some other recent work has developed models which produce natural language explanations for their outputs \cite{park2016attentive,wang2016fvqa}, but there has not been work on generating explanations for irrelevant questions or false premises.

\paragraph{Question Relevance:} Most related to our work is that of \citet{ray2016question}, which introduced the task of irrelevant question detection for VQA. To evaluate on this task, they created the Visual True and False Question (VTFQ) dataset by pairing VQA questions with random VQA images and having human annotators verify whether or not the question was relevant. As a result, many of the irrelevant image-question pairs exhibit a complete mismatch of image and question content. Our Question Relevance Prediction and Explanation (QRPE) dataset on the other hand is collected such that irrelevant images for each question closely resemble the source image both visually and semantically.
We also provide premise-level annotations which can be used to develop models that not only decide whether a question is relevant, but also provide explanations for \emph{why} that is the case.

\paragraph{Semantic Tuple Extraction:} Extracting structured facts in the form of semantic tuples from text is a well studied problem \cite{schuster2015generating,anderson2016spice,elhoseiny2016automatic}; however, recent work has begun extending these techniques to visual domains \cite{xu2017scenegraph,johnson2015image}. Additionally, the Visual Genome \cite{krishnavisualgenome} dataset contains dense image annotations for objects and their attributes and relationships. However, we are the first to consider these facts to reason about question relevancy and compositional reasoning in VQA.
\vspace{\sectionReduceTop}
\section{Extracting Premises of a Question}
\label{sec:premises}
\vspace{\sectionReduceBot}

In \secref{sec:intro}, we introduced the concept of premises and how they can be used. We now formalize this
concept and explain how premises can be extracted from questions. 

We define question premises as
facts implied about an image from a question asked about it, which we represent as tuples. Returning to our
running example question \textit{``What brand of racket is the man holding?''}, we can express these premises
as the tuples \textit{`<man>'}, \textit{`<racket>'}, and \textit{`<man, holding, racket>'} respectively. We categorize these tuples 
into three groups based on their complexity. First-order premises representing the presence of objects
(\textit{`<man>', `<cat>', `<sky>'}), second-order premises capturing the attributes of objects (\textit{`<man, tall>', `<car, moving>'}), and third-order premises containing interactions between objects (\eg 
\textit{`<man, kicking, ball>', `<cat, above, car>'}). 

\begin{figure}
	\centering
	\includegraphics[width=\columnwidth]{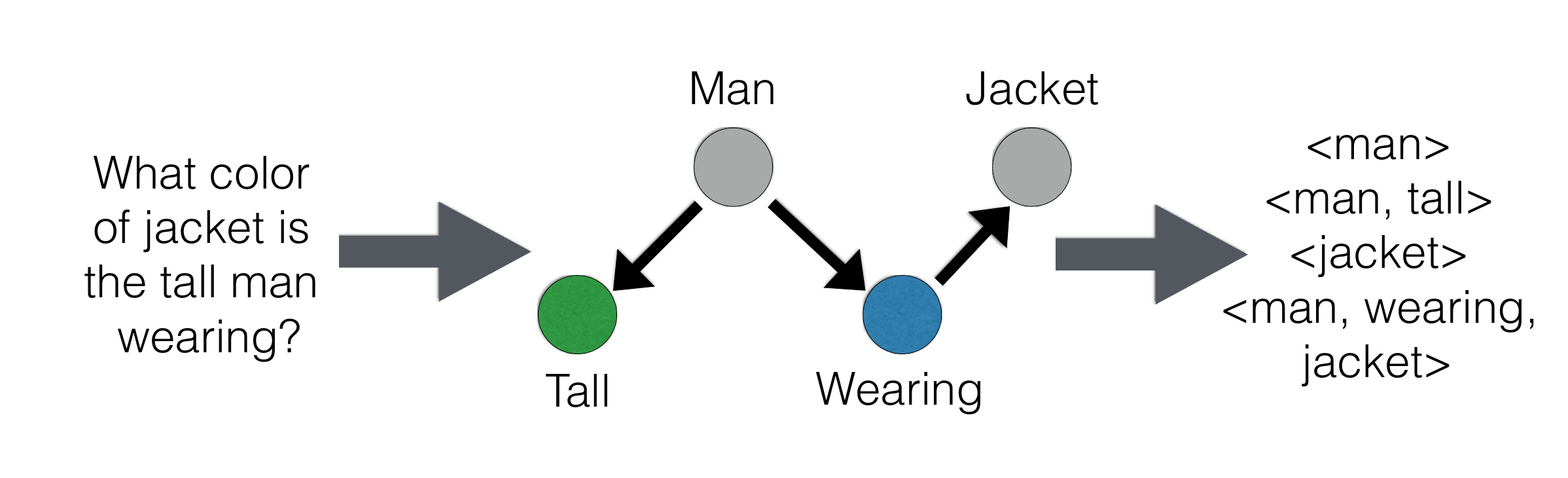}
	\caption{\textbf{Premise Extraction Pipeline.} Objects (gray), attributes (green), and relations (blue) scene graph nodes are converted into 1st, 2nd, and 3rd order premises respectively.}
	\label{fig:prem_ext}
\end{figure}

\paragraph{Premise Extraction:} To extract premises from questions, we use the semantic tuple extraction pipeline used 
in the SPICE metric \cite{anderson2016spice}. Originally defined as a metric for image captioning, SPICE transforms 
a sentence into a scene graph using the Stanford Scene Graph Parser \cite{schuster2015generating} and then extracts
semantic tuples from this representation. \figref{fig:prem_ext} shows this process for a sample question. The question
is represented as a graph of objects, attributes, and relationships from which first, second, and third order premises
are extracted respectively. As this pipeline was originally designed for descriptive captions rather than questions, we
found a number of minor modifications helpful in extracting quality question premises, including disabling pronoun 
resolution, verb lemmatization and METEOR-based Synset matching. We will release our premise extraction code publicly to encourage reproducibility.

While this extraction process typically produces high quality premise tuples, there are some sources of noise which 
must be filtered out. The SPICE process occasionally produces duplicate nodes or object nodes not linked to nouns in
the question, which we filter out. We also remove premises containing words like photo, image, \etc that refer to the 
image rather than its content. 

A more nuanced source of erroneous premises comes from the ambiguity in existential questions, \ie those about the existence of certain image content. For example, while the question 
\textit{``Is the little girl moving?''} contains the premise `\textit{<girl, little>}', it is unclear without the answer whether `\textit{<girl, moving>}' is also
a premise. Similarly, for the question \textit{``How many giraffes are in the image?''}, `\textit{<giraffe, many>}' cannot be considered a premise as there may be 0 giraffes in the image.  
To avoid introducing false premises, we filter out existential and counting questions.
\vspace{\sectionReduceTop}
\section{Question Relevance Prediction and Explanation (QRPE) Dataset}
\vspace{\sectionReduceBot}
\label{sec:qrpe}

\begin{figure*}
\includegraphics[width=\textwidth]{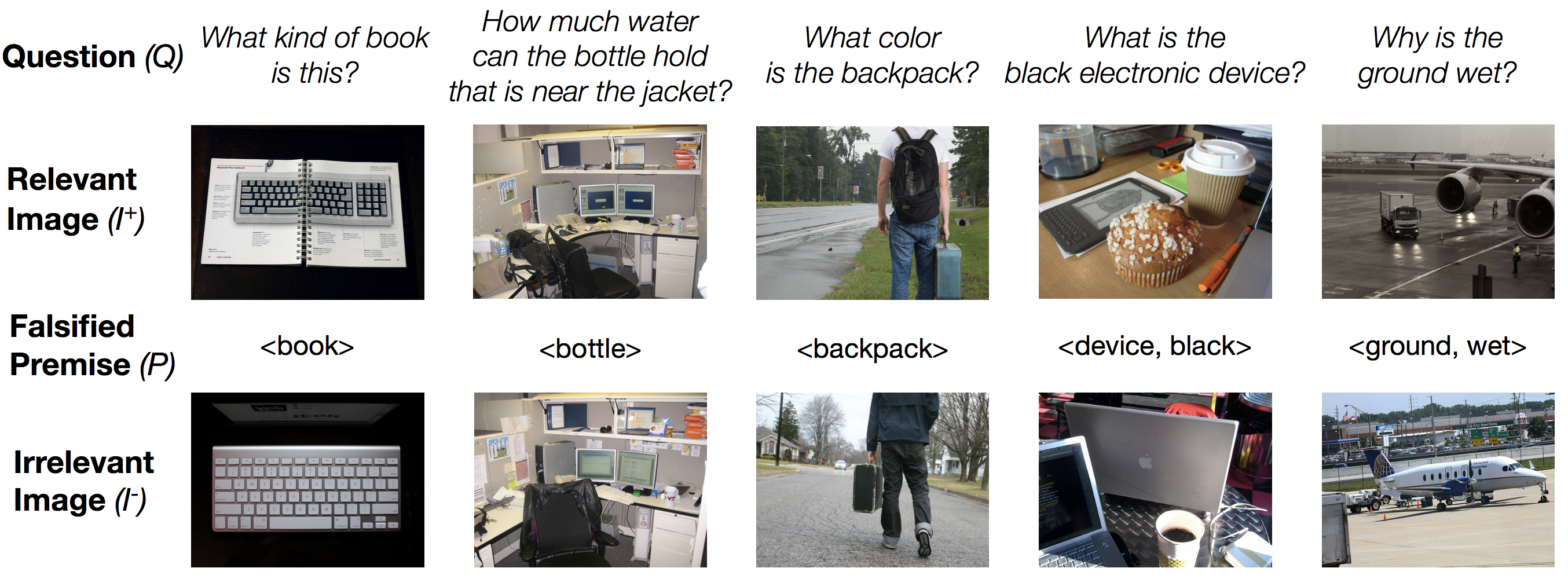}
	\caption{\textbf{Some Examples from QRPE Dataset.} For a given question $Q$ and a relevant image $I^+$, we find an irrelevant image $I^-$ for which exactly one premise $P$ of the question is false. If there are multiple such candidates, we select the candidate most visually most similar to $I^+$. As can be seen from these examples, the QRPE dataset is very challenging, with only minor visual and semantic differences separating the relevant and irrelevant images.}
	\label{fig:qrpe}
\end{figure*}

As discussed in \secref{sec:intro}, modern VQA models fail to differentiate between relevant and irrelevant questions, 
answering either with confidence. This behavior is detrimental to the real world application of VQA systems.
In this section, we curate a new dataset for question relevance in VQA which we call the Question Relevance 
Prediction and Explanation (QRPE) dataset. We plan to release QRPE publicly to help future efforts.

In order to train and evaluate models for irrelevant question detection, we would like to create a dataset of tuples 
$(I^+, Q, P, I^-)$ comprised of a natural language question $Q$, an image $I^+$ for which $Q$ is relevant, and an image 
$I^-$ for which $Q$ is irrelevant because premise $P$ is false. While it is not required to collect both a relevant and irrelevant image for each 
question, we argue that doing so is a simple way to balance the dataset and it ensures that biases against rarer questions
 (which would be irrelevant for most images) cannot be exploited to inflate performance.

We base our dataset on the existing VQA corpus \cite{antol2015vqa}, taking the human-generated (and therefore relevant) 
image-question pairs from VQA as $I^+$ and $Q$. As previously discussed, we can define the relevancy of a question in terms 
of the validity of its premises for an image, so we extract premises from each question $Q$ and must find a suitable irrelevant image $I^-$.
However, there are certainly many images for which one or more of $Q$'s premises are false and an important design decision is 
then how to select $I^-$ from this set.

To ensure our dataset is as realistic and challenging as possible, we consider
 irrelevant images which only have a single false question premise under $Q$ which we denote $P$. For example, the question 
\textit{``Is the big red dog old?''} could be matched with an image containing a big, white dog or a small red dog, but not 
a small white dog. In this way, we ensure that image content is semantically appropriate for the question topic but 
not quite relevant. Additionally, this provides each irrelevant image with an explanation for why the question does not apply.

Furthermore, we sort this subset of irrelevant image by their visual distance to the source image $I^+$ based on image encodings from a VGGNet \cite{Simonyan14c} pretrained on ImageNet \cite{ilsvrc12}. This ensures that
the relevant and irrelevant images are visually similar and act as difficult examples. 

A major difficulty with our proposed data collection process is how to verify whether a premise if true or false for any
given image in order to identify irrelevant images. We detail dataset construction and our approach for this problem in the following section.

\vspace{\subsectionReduceTop}
\subsection{Dataset Construction}
\vspace{\subsectionReduceBot}

\begin{figure*}[t]
	\centering
    \vspace{15pt}
	\includegraphics[width=\textwidth]{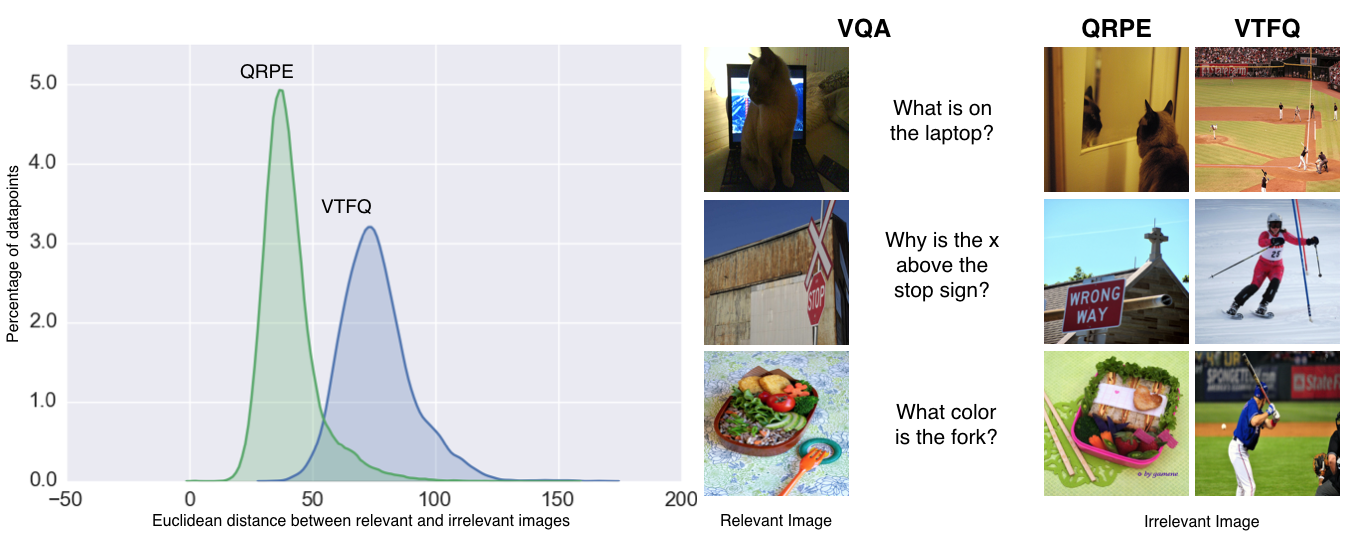}
	\caption{
 	\textbf{A comparison of the QRPE and VTFQ Datasets.} On the left, we plot the Euclidean distance between VGGNet-fc7 features extracted from each relevant-irrelevant image pair for each dataset. Note that VTFQ has significantly higher visual distances. On the right,  we show some qualitative examples of irrelevant images for questions that occur in both datasets. VTFQ images are significantly less related to the source image and question than in our dataset.}
\label{fig:comparison}
\vspace{5pt}
\end{figure*}

We curate our QRPE dataset automatically from existing annotations in COCO \cite{coco2014} and Visual Genome \cite{krishnavisualgenome}.
COCO is a set of over 300,000 images annotated with object segmentations and presence information for 80 classes as well as text descriptions of image content. Visual Genome builds on this dataset, providing more detailed object, attribute, and relationship annotations for over 100,000 COCO images. We make use of these data sources to extract first and second order premises from VQA questions which are also based on COCO images.

For first order premises (\ie existential premises), we consider only the 80 classes present in COCO \cite{coco2014}. As VQA 
and COCO share the same images, we can easily determine if a first order premise is true or false for a candidate irrelevant 
image simply by checking for the absence of the appropriate class annotation. 

For second order premises (\ie attributed objects), we rely on Visual Genome \cite{krishnavisualgenome} annotations for object
and attribute labels. Unlike in COCO, the lack of a particular object label in an image for Visual Genome does not necessarily 
indicate that the object is not present, both due to annotation noise and the use of multiple synonyms for objects by human labelers. 
As a consequence, we restrict the set of candidate irrelevant images to those which contain a matching object to the question premise
but a different attribute. Without further restriction, the selected irrelevant attributes do not tend to be mutually exclusive 
with the source attribute (\ie matching \textit{`<dog, old>'} and \textit{`<dog, red>'}). To correct this and ensure a false premise, we further restrict 
the set to attributes which are antonyms (\eg \textit{`<young>'} for source attribute \textit{`<old>')} or taxonomic sister terms (\eg \textit{`<green>'} for source attribute \textit{`<red>'}) of the original premise attribute. We also experimented with third order 
premises; however, the lack of a corresponding sense of mutual exclusion for verbs and the sparsity of <object, relationship, object> premises made finding non-trivial irrelevant images difficult.

To recap, our data collection approach is to take each image-question pair in the VQA dataset and extract its first and second order 
question premises. For each premise, we find all images which lack only this premise and rank them by their visual distance. The
closest of these is kept as the irrelevant image for each image-question pair. 

\vspace{\subsectionReduceTop}
\subsection{Exploring the Dataset}
\vspace{\subsectionReduceBot}

\figref{fig:qrpe} shows sample $(I^+, Q, P, I^-)$ tuples from our dataset. These examples illustrate the difficulty of our dataset. For instance, the images in the second column
differ only by the presence of the water bottle and images in the fourth column are differentiated by the color of the devices. Both of these are fine details of the image content.

The QRPE dataset contains {53,911} $(I^+, Q, P, I^-)$ tuples generated from as many premises. In total, it contains {1530} unique premises and {28,853} unique questions. Among the {53,911} premises, {3876} are second-order, attributed object premises while the remaining {50,035} are first-order object/scene premises. We divide our dataset into two parts -- a training set with {35,486} tuples that are generated from the VQA training set and a validation set with {18,425} tuples generated from the VQA validation set.

\paragraph{Manual Validation.} We also manually validated 1000 randomly selected $(I^+, Q, P, I^-)$ tuples from our dataset. We noted that 99.10\% of the premises $P$ were valid (\ie implied by the question) in $I^+$ and 97.3\% were false for the negative image $I^-$. This demonstrates the high reliability of our automated annotation pipeline.

\vspace{\subsectionReduceTop}
\subsection{Comparison to VTFQ}
\vspace{\subsectionReduceBot}

We contrast our approach to the VTFQ dataset of \citet{ray2016question}. As discussed prior, VTFQ was collected by selecting a random question
and image from the VQA set and asking human annotators to report if the question was relevant, producing a pair. This approach results in irrelevant 
image-question pairs that are unambiguously unrelated, with the visual content of the image having nothing at all to do with
the question or its source image from VQA. 

\sml{To quantify this effect and compare to QRPE, we pair each irrelevant image-question pair $(I^-,Q)$ from VTFQ with a relevant image from the VQA dataset. Specifically, we find the nearest neighbor question $Q^{nn}$ in the VQA dataset to $Q$ based on an average of the word2vec~\cite{mikolov2013distributed} embedding of each word, and select the image on which $Q^{nn}$ was asked as $I^+$ to form $(I^+, Q, P, I^-)$ tuples like in our proposed dataset.}

\sml{In \figref{fig:comparison}, we present a quantitative and qualitative comparison of the two datasets based on these tuples. On the left side of the figure, we plot the distributions of Euclidean distance between the fc7 features of each $(I^+,I^-)$ pair in both datasets. We find that the mean distance in the VTFQ dataset is nearly twice that of our QRPE dataset, indicating that irrelevant images in VTFQ are less visually related to source images though we do note the distribution of distances in both datasets is long tailed.}

On the right side of \figref{fig:comparison}, we also provide qualitative examples of questions that occur in both datasets. The example on the last row is perhaps most striking. The source question is asking the color of a fork and the relevant image shows an overhead view of a meal with an orange fork set nearby. The irrelevant image in QRPE is a similar image of food, but with chopsticks! Conversely, the image from VTFQ is a man playing baseball.

\vspace{\sectionReduceTop}
\section{Question Relevance Detection}
\label{sec:qr_models}
\vspace{\sectionReduceBot}

In this section, we introduce a simple baseline for irrelevant question detection on the QRPE dataset and demonstrate that explicitly reasoning about premises improves performance for both our new model and existing methods. More formally, we consider the binary classification task of predicting if a question $Q_i$ from an image-question pair $(I_i,Q_i)$ is relevant to image $I_i$.

\paragraph{A Simple Premise-Aware Model.} Like the standard VQA task, question relevance detection also requires making a 
prediction based on an encoded image and question. With this in mind, we begin with a straight-forward approach based 
on the Deeper LSTM VQA model architecture of \citet{antol2015vqa}. This model encodes the image $I$ via a VGGNet and the 
question $Q$ with an LSTM over one-hot word encodings. The concatenation of these embeddings are input to a multi-layer perceptron. We 
fine-tune this model for the binary question relevance detection task starting from a model pre-trained on the VQA task. We denote 
this model as \texttt{VQA-Bin}. 

We extend the \texttt{VQA-Bin} model to explicitly reason about premises. We extract first and second order premises from the question $Q$ and encode them as two concatenated one-hot vectors. We add an additional LSTM to encode the premises and concatenate this added feature to the
image and question feature. We refer to this premise-aware model as \texttt{VQA-Bin-Premise}.

\paragraph{Attention Models.} We also extend the attention based Hierarchical Co-Attention VQA model of \citet{lu2016hierarchical} for the task of question relevance in a way similar to Deeper LSTM model. We call this model \texttt{HieCoAtt-Bin}. The corresponding premise-aware model is referred to as \texttt{HieCoAtt-Bin-Prem}.

\paragraph{Existing Methods.} We compare our approaches with the best performing model of \citet{ray2016question}. 
This model (which we denote \texttt{QC-Sim}) uses a pretrained captioning model to automatically provide natural language  image descriptions and reasons about relevance based on a learned similarity between the question and image caption. 

\begin{figure*}[t]
 	\centering
 	\includegraphics[width= \linewidth]{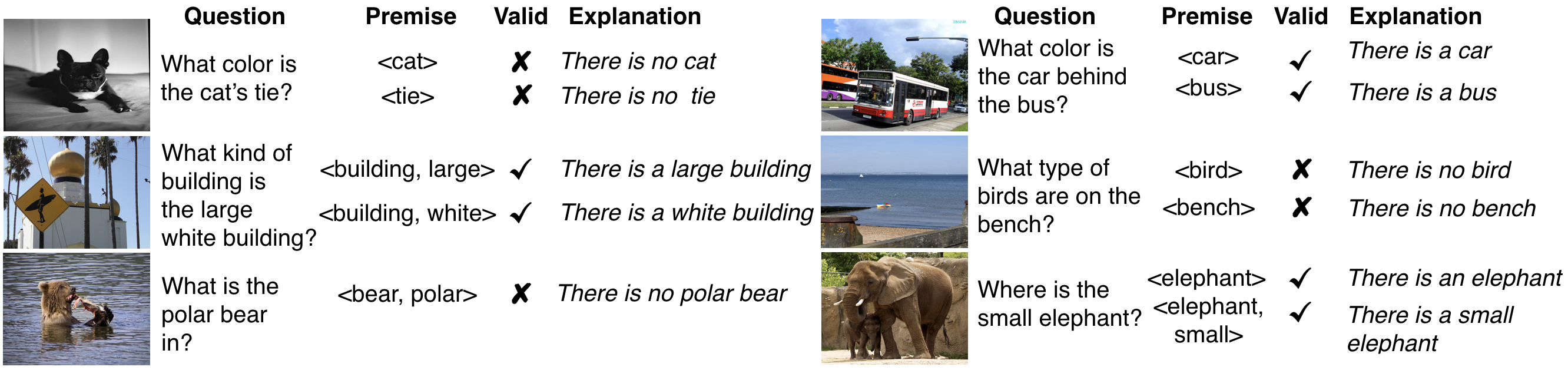}
 	\caption{
 	\textbf{Question relevance explanation:} We provide selected examples of predictions from the False Premise Detection model (\texttt{FPD}) on the QRPE test set. Reasoning about premises presents the opportunity to produce natural language statements indicating \textit{why} a question is irrelevant to an image, by pointing to the premise that is invalid.
 	}
	\label{fig:explanation}
    \vspace{0pt}
 \end{figure*}

\sml{Specifically, the approach uses 
NeuralTalk2 \cite{karpathy15cvpr} trained on the MS COCO dataset \cite{coco2014} to generate a caption for each image.
Both the caption and question are embedded as a fixed length vector through an encoding LSTM (with words being represented as word2vec \cite{mikolov2013distributed} vectors).
These question and caption embeddings are concatenated and fed to a multilayer perceptron to predict relevance. We consider two additional versions of this approach that consider only premise-caption similarity (\texttt{PC-Sim}) and question-premise-caption similarities (\texttt{QPC-Sim}).}

\paragraph{Results.} We train each model on the QRPE train split and report results on the test set in \tabref{tab:fp_models}. {As the dataset is balanced in the label space, random accuracy stands at 50\%}. We find that the simple \texttt{VQA-Bin} model achieves {66.5\%} accuracy {while the attention based model \texttt{HieCoAtt-Bin} attains 70.74\% accuracy.} Surprisingly, the caption-similarity based \texttt{QC-Sim} model significantly outperforms these baseline, obtaining an accuracy of {74.35\%} \sml{while only reasoning about relevancy from textual descriptions of images. We note that the caption similarity based approaches use a large amount of outside data during pretraining of the captioning model and the word2vec embeddings, which may have contributed to the effectiveness of these methods.}

\begin{table}
	\centering
	\footnotesize
	\setlength{\tabcolsep}{8.5pt}
    \resizebox{\columnwidth}{!}{
		\begin{tabular}{ c c c c}
			\toprule
			Models & Overall & First Order & Second Order\\
			\cmidrule{1-4}
            VQA-Bin & 66.50 & 67.36 & 53.00\\
            VQA-Bin-Prem&66.77 &67.04 & 54.38\\
            \cmidrule{1-4}
            {HieCoAtt-Bin} & 70.74 & 71.35 & \textbf{61.54}\\
            {HieCoAtt-Bin-Prem}& 73.34 &73.97 &60.35\\
            \cmidrule{1-4}
            QC-Sim& 74.35 &75.82 &55.12\\
            PC-Sim& 75.05&76.47& 56.04\\
            QPC-Sim& \textbf{75.31}&\textbf{76.67}&55.95\\
			\bottomrule
		\end{tabular} }
		\caption{Accuracy of  Question Relevance models on the QRPE test set. We find that premise-aware models consistently outperform alternative models. \label{tab:fp_models}}
\end{table}

Most interestingly, we find that the addition of extracted premise representations consistently improves performance of base models. {\texttt{VQA-Bin-Prem}, \texttt{HieCoAtt-Bin-Prem}, \texttt{PC-Sim}, and \texttt{QPC-Sim}} outperform their no-premise information counterparts, with \texttt{QPC-Sim} being the overall best performing approach at {75.31}\% accuracy. This is especially interesting given that the models \emph{already} have access to the question from which the premises were extracted. This result seems to imply there is value in explicitly isolating premises from sentence grammar.

We further divide our test set into two splits consisting of $(Q, I)$ pairs created by either falsifying first-order and second-order premises. We find that all our models perform significantly better on the first-order split. We hypothesize that the significant diversity in visual representations of attributed objects and comparatively fewer examples for each type makes it more difficult to learn subtle differences for second-order premises.

\vspace{\subsectionReduceTop}
\subsection{Question Relevance Explanation}
\vspace{\subsectionReduceBot}

In addition to identifying whether a question is irrelevant to an image, being able to indicate \emph{why} carries significant real-world utility. From an interpretability perspective, reporting which premise is false is more informative than simply answering the question in the negative,  
as it can help to correct the questioner's misconception regarding the scene. We propose to generate such explanations by identifying the particular question premise(s) that do not apply to an image.

By construction, irrelevant images in the QRPE dataset are picked on the basis of negating a single premise -- we now use our dataset to train models to detect false premises, and use the premises classified as irrelevant to generate templated natural language explanations.

\figref{fig:explanation} illustrates the task setup for false premise detection. Given a question-image pair, say \textit{``What color is the cat's tie?''}, the objective is to identify which (if any) question premises are not grounded in the image, in this case both \textit{<cat>} and \textit{<tie>}. Alternatively, for the question \textit{``What kind of building is the large white building?''}, both premises \textit{<building, large>} and \textit{<building, white>} are true premises grounded in the image.

\newcolumntype{L}{>{\centering\arraybackslash}m{0.38\textwidth}}
 \begin{figure*}
 \centering
 \def\arraystretch{1.05}
 \resizebox{0.95\textwidth}{!}{
 \begin{tabular}{L L L}
 \large\textbf{What player number is about to swing at the ball?} &
 \large\textbf{Why is the man looking at the lady?} &
 \large\textbf{How many people are wearing safety jackets?}\\
 Is there a player number? Yes & 
 Who is looking at the lady? Man&
 Can you see people in the image? Yes\\
 Is there a ball in the image? Yes &
 Is there a lady in the image? Yes &
 What are the people wearing? Jacket\\
 Is there a number in the image? Yes &
 Is there a man in the image? Yes & 
 Who is wearing the jacket? People\\
 \vspace{0pt}\large\textbf{What is the child sitting on?}&
 \vspace{0pt}\large\textbf{Where is the pink hat?}&
 \vspace{11pt}\large\textbf{What is the item called that the cat is looking at?}\\
 What is the child doing? Sitting&
 What is the color of hat? Pink &
 Is there a cat in the image? Yes\\
 Is there a child in the image? Yes&
 Is there a hat in the image? Yes&
 Is there an item in the image? Yes\\
 & 
 \end{tabular}}
 \caption{Sample generated premise questions from source questions. Source questions are in bold. Ground-truth answers are extracted using the premise tuples. \label{fig:sampleq}}
 \vspace{0pt}
 \end{figure*}

We train a simple false premise detection model for this task. Our model is a multilayer perceptron that takes one-hot encodings of premises and VGGNet \cite{Simonyan14c} image features as input to predict whether the premise is grounded in the image or not.
We trained our false premise detection model (\texttt{FPD}) model on all premises in the QRPE dataset.

\par
Our \texttt{FPD} model achieves an accuracy of {61.12}\% on the QRPE dataset. {In \figref{fig:explanation}, we present qualitative results of our premise classification and explanation pipeline. For the question \textit{``What color is the cat's tie?''}, the model correctly recognizes `cat' and `tie' as false premises, and we generate statements in natural language indicating the same. Thus, determining question relevance by reasoning about each premise presents the opportunity to generate simple explanations that can provide valuable feedback to the questioner, and help improve model trust.
\vspace{\sectionReduceTop}
\section{\sml{Premise-Based Visual Question Answering Data Augmentation}}
\label{sec:augment}
\vspace{\sectionReduceBot}

In this section, we develop a premise-based data augmentation scheme for VQA that generates simple, templated questions 
based on premises present in complex visually-grounded questions from the VQA (training) dataset. 

Using the pipeline presented in \secref{sec:premises}, we extract premises from questions in the VQA dataset and apply
a simple templated question generation strategy to transform premises into question and answer pairs. Note that because the source questions come from sighted humans about an image, we do not need to filter out binary or counting questions in order to avoid false premises as in \secref{sec:premises}. We do however filter based on SPICE similarity between the generated and source questions to avoid generating duplicates.

We design templates 
for each type of premise -- first-order (\eg \textit{`<man>'} -- \textit{``Is there a man?'' Yes}), second-order 
(\eg \textit{`<man, walking>'} -- \textit{``What is the man doing?'' Walking}, and \textit{`<car, red>'} -- \textit{``What is 
the color of the car?'' Red}), and third-order (\textit{`<man, holding, racket>'} -- \textit{``What is the man holding?'' Racket, 
``Who is holding the racket?'' Man}). This process transforms implicit premise concepts which previously had to be learned as part of understanding more complex questions into simple, explicit training examples that can be directly supervised.

\figref{fig:sampleq} shows sample premise questions produced from source VQA questions using our pipeline. We note that the distribution of premise questions varies drastically from the source VQA distribution (see \tabref{tab:comp_data}). 

\begin{table} 
    \centering
    \footnotesize
	\setlength{\tabcolsep}{2 pt}
	\resizebox{\columnwidth}{!}{
		\minipage{\columnwidth}
		\begin{tabular}{@{} l  c  c  c  c  c  c  c @{}}
			\toprule
			Training Data    & Other   & Number & Yes &No  & Total\\ 
			\midrule
			Source& 123,817 & 29,698 & 57217 & 35842 & 246,574 \\
			Premise & 137,483 & 1,850  & 387,941 & 0  & 527,274\\
		 \bottomrule
		\end{tabular}
		\captionof{table}{\label{tab:comp_data} Answer type distribution of source and premise questions on the Compositional VQA train set.}
		\endminipage }
	\vspace{0pt}
\end{table}

We evaluate multiple models with and without premise augmentation on two splits of the VQA dataset - the standard split and the compositional split of \citet{c_vqa}. The compositional split is specifically designed to test a model's ability to generalize to unseen/rarely seen combinations of concepts at test time.

\paragraph{Augmentation Strategies.} We evaluate the Deeper LSTM model of \citet{Lu2015} 
on the standard and compositional splits with two augmentation strategies - \texttt{All} which includes the entire set of premise questions and \texttt{Top-1k-A} which includes only questions with answers in the top 1000 most common VQA answers. The results are listed in \tabref{tab:vqa_comp}. We find minor improvement of 0.34\% on the standard split under \texttt{Top-1k-A} premise question augmentation. On the compositional split, we observe a 1.16\% gain with \texttt{Top-1k-A} augmentation over no augmentation. In this setting, explicitly reasoning about objects and attributes seen in the questions seems to help the model disentangle objects from their common characteristics.

\paragraph{Other Models.} To check the general effectiveness of our approach, we further evaluate \texttt{Top-1k-A} augmentation for three additional VQA models on the compositional split. We find inconsistent improvements for these more advanced models with some improving while others see reductions in accuracy when adding premises.

\begin{table} 
	\centering
	\footnotesize
    \renewcommand*{\arraystretch}{1.25}
	\setlength{\tabcolsep}{3pt}
        \resizebox{\columnwidth}{!}{
		\begin{tabular}{@{} l l  c  c  c  c  c  c  c @{}}
       		\toprule
			& Augmentation & Overall & Other & Number & Yes/No\\
			\midrule
			\multirow{3}{*}{\rotatebox{90}{Standard}}& None  & 54.23 & 40.34 & 33.27 & 79.82\\
			& All & 53.74 & 39.28 & \textbf{33.38} & 79.89\\
			& Top-1k-A & \textbf{54.47} & \textbf{40.56} & 33.24 & \textbf{80.19}\\
			\midrule
			\multirow{3}{*}{\rotatebox{90}{Comp.}} & None & 46.69 & 31.92 & 29.73 & 70.49\\
			& All & 47.63 & 31.97 & \textbf{30.77} & \textbf{72.52}\\
			& Top-1k-A& \textbf{47.85} & \textbf{32.58} & 30.59 & 72.38\\
			\bottomrule
		\end{tabular}}

		\caption{\label{tab:vqa_comp}Accuracy on the standard and compositional VQA validation sets for different augmentation strategies for DeeperLSTM\cite{antol2015vqa}.}
        \vspace{10pt}
\end{table}

\begin{table}
	\footnotesize
    \renewcommand*{\arraystretch}{1.25}
	\setlength{\tabcolsep}{2 pt}
	\resizebox{\columnwidth}{!}{
		\centering
    	\begin{tabular}{@{} l  c  c  c  c  c  c  c @{}}
			\toprule
			VQA Model & Baseline & +Premises&\\
			\midrule
			DeeperLSTM\cite{Lu2015} &46.69	& \textbf{47.85}\\
			HieCoAtt\cite{lu2016hierarchical} &\textbf{50.17}	&49.98	\\
			NMN\cite{andreas2016neural} &\textbf{49.05}&48.43\\
			MCB\cite{fukui2016multimodal} & 50.13&\textbf{50.57}\\
			\bottomrule
		\end{tabular}
        }
		\caption{\label{tab:comp_other}Overall accuracy of different VQA models on the Compositional VQA test split using Top-1k-A augmentation.}
\end{table}
\section{Conclusions and Future Work}
\label{sec:conclusion}
\vspace{\sectionReduceBot}
In this paper, we made the simple observation that questions about images often contain premises implied by the question and that reasoning about premises can help VQA models respond more intelligently to irrelevant or novel questions. 

We develop a system for automatically extracting these question premises. Using these premises, we automatically created a novel dataset for Question Relevance Prediction and Explanation (QRPE) which consists of 53,911 question, relevant image, and irrelevant image triplets. We also train novel question relevance prediction models and show that models that take advantage of premise information outperform models that do not. Furthermore, we demonstrated that questions generated from premises may be an effective data augmentation technique for VQA tasks that require compositional reasoning.

Integrating Question Relevance Prediction and Explanation (QRPE) models with existing VQA systems would form a natural extension to our approach. In this setting, the relevance prediction model would determine the applicability of a question to an image, and select an appropriate path of action. If the question is classified as relevant, the VQA model would generate a prediction; otherwise, a question relevance explanation model would provide a natural language sentence indicating which premise(s) are not valid for the image. Such systems would be a step in the direction of making VQA systems move beyond academic settings to real-world environments.
\bibliography{emnlp2017}
\bibliographystyle{emnlp_natbib}

\clearpage
\appendix
\vspace{\sectionReduceTop}
\section*{\LARGE Appendix}
\vspace{\sectionReduceBot}
\label{sec:appendix}
\vspace{\sectionReduceTop}
\section{Compositional VQA Split}
\label{sec:dataset}
\vspace{\sectionReduceBot}

In this section, we provide details regarding the Compositional VQA split introduced by \citet{c_vqa}, on which we perform our data augmentation experiments (\secref{sec:augment}).
The compositional splits were created by re-arranging the training and validation splits of the VQA dataset \citep{antol2015vqa}. These splits were created such that the question-answer (QA) pairs in the compositional test split (\eg, Question: ``What color is the plate?'', Answer: ``green'') are not seen in the compositional train split, but the concepts that compose the test QA pairs (\eg, ``plate'', ``green'') have been seen in the compositional train split (\eg, Question: ``What color is the apple?'', Answer: ``Green'', Question: ``How many plates are on the table?'', Answer: ``4'') to the extent possible. Evaluating a VQA model under such a setting helps in testing -- 1) whether the model is capable of learning disentangled representations for different concepts (\eg, ``plate'', ``green'', ``apple'', ``4'', ``table''), and 2) whether the model can compose these learned concepts to correctly answer questions about novel questions at test time.

\vspace{\sectionReduceTop}
\section{Question Generation}
\label{sec:generation}
\vspace{\sectionReduceBot}

 \newcolumntype{L}{>{\centering\arraybackslash}m{0.38\textwidth}}
 \begin{figure*}
 \centering
 \def\arraystretch{1.05}
 \resizebox{0.95\textwidth}{!}{
 \begin{tabular}{L L L}
 \large\textbf{What player number is about to swing at the ball?} &
 \large\textbf{Why is the man looking at the lady?} &
 \large\textbf{How many people are wearing safety jackets?}\\
 Is there a player number? Yes & 
 Who is looking at the lady? Man&
 Can you see people in the image? Yes\\
 Is there a ball in the image? Yes &
 Is there a lady in the image? Yes &
 What are the people wearing? Jacket\\
 Is there a number in the image? Yes &
 Is there a man in the image? Yes & 
 Who is wearing the jacket? People\\
 \vspace{0pt}\large\textbf{What is the child sitting on?}&
 \vspace{0pt}\large\textbf{Where is the pink hat?}&
 \vspace{11pt}\large\textbf{What is the item called that the cat is looking at?}\\
 What is the child doing? Sitting&
 What is the color of hat? Pink &
 Is there a cat in the image? Yes\\
 Is there a child in the image? Yes&
 Is there a hat in the image? Yes&
 Is there an item in the image? Yes\\
 & 
 \end{tabular}}
 \caption{Sample generated premise questions from source questions. Source questions are in bold. Ground-truth answers are extracted using the premise tuples. \label{fig:sampleq}}
 \vspace{0pt}
 \end{figure*}
 
 \begin{figure*}
	\centering
	\includegraphics[width=\textwidth]{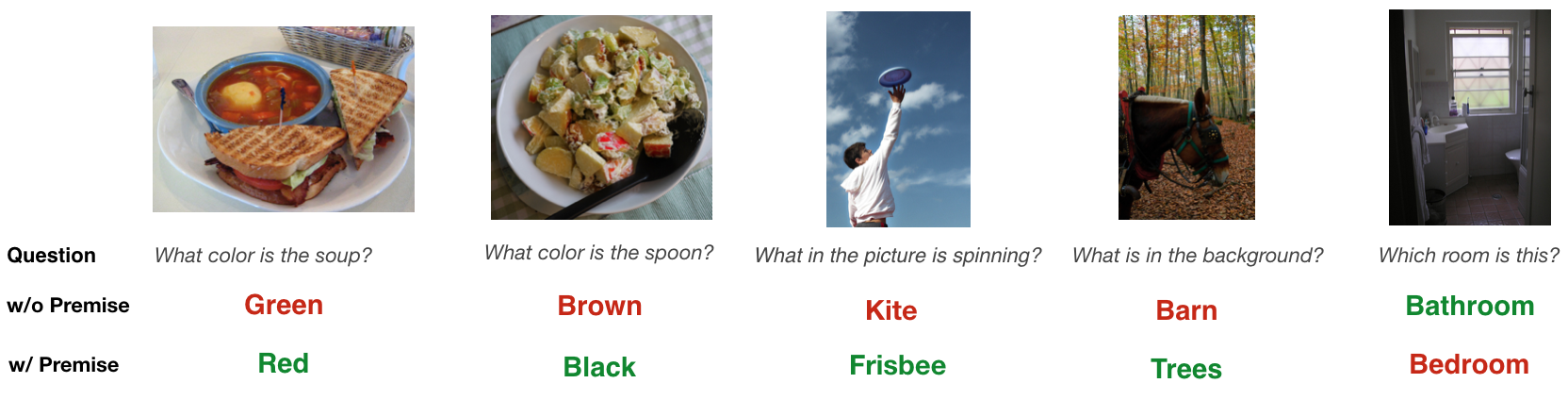}
	\caption{Some interesting examples of how augmentation helps the DeeperLSTM model \citep{antol2015vqa} on the compositional VQA split.}
\label{fig:compositionality}
\vspace{0pt}
\end{figure*}

For the data augmentation experiments in \secref{sec:augment}, we generate premise questions using a rule-based pipeline.  Different templates of questions are assigned for different kinds of facts.

First order premises like <man>, <bus> are transformed into existential questions like ``Is there a man?'', ``Is there a bus?'' and so on. Second order premises can generate two kinds of questions depending on whether the second element is an action or an attribute. For example, <man, walking> would become ``What is the man doing?'' while <car, red> would become "What is the color of the car?''. In general, questions generated from third order premises look like ``Is the man holding the racket?", and ``What is the cat on top of?'' for <man, holding, racket> and <cat, on top of, box>, respectively. However, third order premises are more complicated and many different questions can be generated from them depending on the types of components in the premise.

Question generation also involves minor pre-processing and post-processing, i.e. filtering out erroneous premises and linguistically ambiguous questions. We also run SPICE on the generated and source questions and threshold the result to eliminate generated questions that are near duplicates of the source questions. Code for our question generation pipeline will be made available.

A random selection of premise questions generated from the VQA dataset can be seen in \figref{fig:sampleq}. The answer type distribution of generated premise questions can be seen in \tabref{tab:comp_data}. We find that generated premise questions are twice in number as compared to source questions. We generate relatively few `Number' questions -- very few second-order tuples of this type occur in the premises we extract, as questions about multiple number of objects at a time are rare in the VQA dataset. By design, we generate only `Yes' questions and zero `No' questions. The reason for that is twofold -- first, we only generate premise questions from true premises, and second, first order premises are the most frequent premises in source questions (first order premises generate `Yes' questions).

\begin{table} 
	\footnotesize
	\setlength{\tabcolsep}{2 pt}
	\resizebox{\columnwidth}{!}{
		\minipage{\columnwidth}
		\begin{tabular}{@{} l  c  c  c  c  c  c  c @{}}
			\toprule
			Training Data    & Other   & Number & Yes &No  & Total\\ 
			\midrule
			Source& 123,817 & 29,698 & 57217 & 35842 & 246,574 \\
			Premise & 137,483 & 1,850  & 387,941 & 0  & 527,274\\
		 \bottomrule
		\end{tabular}
		\captionof{table}{\label{tab:comp_data} Answer type distribution of source and premise questions on the Compositional VQA train set.}
		\endminipage }
	\vspace{0pt}
\end{table}

\vspace{\subsectionReduceTop}
\subsection{Data Augmentation}
\label{sec:dataaug}
\vspace{\subsectionReduceBot}

We perform a series of data augmentation experiments using the questions generated in \ref{sec:generation} and evaluate performance of models on both the standard VQA split and the Compostitional VQA split described in \ref{sec:dataset}. 

\vspace{\subsectionReduceBot}
\subsection{Experimental Setup}
\vspace{\subsectionReduceBot}
For the augmentation experiments, we start by generating premise questions from the Compositional VQA train split \cite{c_vqa}. The generated premise questions are added to the original source questions for training models. The number of generated premise questions is almost twice the number of source questions, therefore we try a series of augmentation strategies based on different subsets of premise questions to be added. The model used for these experiments is the DeeperLSTM model by \cite{antol2015vqa}. The various augmentation ablations are:
\begin{compactenum}[\hspace{0pt}-]
	\item \textbf{Baseline:} No premise questions added to the training set. 
	\item \textbf{All:} Adding all the generated premise questions along with source questions to the training set.
	\item \textbf{Only-Binary:} Only binary (Questions with answers `Yes' or `No') premise questions added along with the source questions.
	\item \textbf{No-Other:} All questions except premise questions of type `Other' (answers outside of Binary and Number answers) added to the training set.
	\item \textbf{No-Binary:} All questions except binary premise questions are added to the training set.
	\item \textbf{Comm-Other:} All binary premise questions added. `Other' and `Number' premise question types whose answers lie in the pool of source question answers are added to the training set.
	\item \textbf{Top1k-A:} All binary premise questions added. Also, premise questions of type `Other' with answers amongst the top 1000 VQA source answers are added. 
\end{compactenum}

\vspace{\subsectionReduceBot}
\subsection{Analysis and Results}
\vspace{\subsectionReduceBot}

\begin{table}
	\centering
	\footnotesize
	\setlength{\tabcolsep}{3pt}
        \resizebox{0.9\columnwidth}{!}{
		\begin{tabular}{@{} l  c  c  c  c  c  c  c @{}}
			\toprule
			Data Ablation & Overall & Other & Number & Yes/No\\
			\midrule
			Baseline & 46.69 & 31.92 & 29.73 & 70.49\\
			All & 47.63 & 31.97 & \textbf{30.77} & \textbf{72.52}\\
			Only-Binary & 47.25 & 32.45 & 29.65 & 71.30\\
			No-Other & 47.33 & 32.47 & 29.85 & 71.42\\
			No-Binary & 46.76 & 31.69 & 29.39 & 71.09\\
			Comm-Other& 47.53 & 32.41 & 28.88 & 72.33\\
			Top1k-A & \textbf{47.85} & \textbf{32.58} & 30.59 & 72.38\\
			\bottomrule
		\end{tabular}}
		\caption{\label{tab:vqa_comp}Performance of DeeperLSTM \cite{antol2015vqa} on Compositional VQA test split with different augmentations.}
\vspace{-10pt}
\end{table}

\begin{table}
	\footnotesize
	\setlength{\tabcolsep}{2 pt}
	\resizebox{\columnwidth}{!}{
		\minipage{\columnwidth}
		\centering
    	\begin{tabular}{@{} l  c  c  c  c  c  c  c @{}}
			\toprule
			VQA Model & Baseline & With Premises&\\
			\midrule
			DeeperLSTM\cite{antol2015vqa} &46.69	& 47.85\\
			HieCoAtt\cite{lu2016hierarchical} &50.17	&49.98	\\
			NMN\cite{andreas2016neural} &49.05&48.43\\
			MCB\cite{fukui2016multimodal} & 50.13&50.57\\
			\bottomrule
		\end{tabular}
		\caption{\label{tab:comp_other}Accuracy of different VQA models on the Compositional VQA test split using Top1k-A augmentation.}
		\endminipage
	}
		\vspace{10pt}
\end{table}

\tabref{tab:vqa_comp} shows the VQA accuracy of the DeeperLSTM model for these different dataset augmentation strategies. While all settings show some improvements over the standard training set, we find the largest increase with the \texttt{Top1k-A} setting. By restricting the additional question to those having answers in the top-1000 most commonly occurring answers from the standard VQA set, the added data does not significantly shift the types of answers the model learns are likely. Some examples where the augmented DeeperLSTM model performs better than a non-augmented model are shown in \figref{fig:compositionality}.

Keeping the \texttt{Top1k-A} data augmentation setting, we extend our experiments to additional VQA models. \tabref{tab:comp_other} shows the results  of these experiments. We find that while this data augmentation technique results in improvements for some models, it fails to consistently deliver significantly better performance overall. 

\end{document}